\documentclass[letterpaper, 10 pt, conference]{ieeeconf}  

\IEEEoverridecommandlockouts                              

\usepackage{graphicx} 
\usepackage{amsmath} 
\usepackage{amssymb}  
\usepackage{float}
\usepackage{xcolor}
\usepackage[caption=false, font=footnotesize]{subfig} 
\let\labelindent\relax
\usepackage{enumitem}
\usepackage{siunitx}

\usepackage[font=footnotesize]{caption}
\usepackage{titlesec}

\usepackage{tabularx,booktabs}
\usepackage{array}
\newcolumntype{P}[1]{>{\centering\arraybackslash}p{#1}}
\newcolumntype{M}[1]{>{\centering\arraybackslash}m{#1}}

\titlespacing{\section}{0pt}{*0.5}{*0.5}
\titlespacing{\subsection}{4pt}{*0.5}{*0.5}
\usepackage{todonotes}

\title{\LARGE \bf
Learning When to Switch: Composing Controllers to Traverse a Sequence of Terrain Artifacts
}

\author{Brendan Tidd$^{1,4}$, Akansel Cosgun$^{2}$, J\"{u}rgen Leitner$^{1,3}$ and Nicolas Hudson$^{4}$
\thanks{$^{1}$Queensland University of Technology (QUT), Australia.}%
\thanks{$^{2}$Monash University, Australia}%
\thanks{$^{3}$LYRO Robotics Pty Ltd, Australia}%
\thanks{$^{4}$Robotics and Autonomous Systems Group, CSIRO, Pullenvale, QLD 4069, Australia}%
}

\begin{document}

\maketitle
\thispagestyle{empty}
\pagestyle{empty}

\begin{abstract}

Legged robots often use separate control policies that are highly engineered for traversing difficult terrain such as stairs, gaps, and steps, where switching between policies is only possible when the robot is in a region that is common to adjacent controllers. Deep Reinforcement Learning (DRL) is a promising alternative to hand-crafted control design, though typically requires the full set of test conditions to be known before training. DRL policies can result in complex (often unrealistic) behaviours that have few or no overlapping regions between adjacent policies, making it difficult to switch behaviours. In this work we develop multiple DRL policies with Curriculum Learning (CL), each that can traverse a single respective terrain condition, while ensuring an overlap between policies. We then train a network for each destination policy that estimates the likelihood of successfully switching from any other policy. We evaluate our switching method on a previously unseen combination of terrain artifacts and show that it performs better than heuristic methods. While our method is trained on individual terrain types, it performs comparably to a Deep Q Network trained on the full set of terrain conditions. This approach allows the development of separate policies in constrained conditions with embedded prior knowledge about each behaviour, that is scalable to any number of behaviours, and prepares DRL methods for applications in the real world.

\end{abstract}

\section{Introduction}

Legged robots are useful for traversing various terrain conditions where wheeled platforms fail to operate. Terrains that are easily negotiated by humans, however, can present a difficult control problem in robotics \cite{atkeson2016happened}. Consider a delivery driver who jumps down from the delivery vehicle, takes a large step over a break in the sidewalk and walks up the stairs to reach the receiver’s front door. For a last mile delivery robot, it is difficult to design a single locomotion controller that can handle a set of similar tasks. Humanoid robots often use separate control policies that have been meticulously tuned for a specific condition in a constrained setting. 


\begin{figure}[tb!]
\centering
\includegraphics[width=1.0\columnwidth]{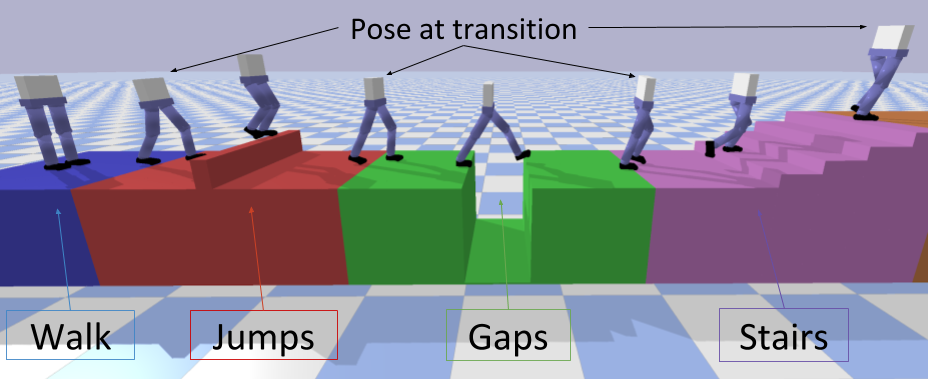}
\caption{We design policies for each complex terrain type separately, ensuring common regions of the state between policies. Switching in these common regions results in desirable behaviour from the switched policy. To successfully switch between policies we train a switch estimator for each policy that learns when best to switch. Our method involves training on each artifact type separately, where training on multiple artifact types at once may be dangerous or implausible.}
\label{fig:main_pic}
\vspace{-3mm}
\end{figure}

For highly dynamic robots such as bipeds, transitioning from one policy to the next may only be possible through a very narrow set of states. Activating a policy when the robot is in an unsuitable configuration for the target policy can result in the robot falling over. We refer to the set of all states from which using a policy will result in a desirable stable behaviour as the Region of Attraction (RoA) of the policy~\cite{burridge_sequential_1999}. A state that is in the RoA for more than one policy presents an opportunity where these policies can be switched such that the agent will eventuate in a stable configuration. A robust policy switching method would reduce robot falls, which is a real problem with humanoid robots \cite{atkeson2016happened,kalyanakrishnan2011learning}.

Deep Reinforcement Learning (DRL) has been shown to be an effective method for developing walking policies~\cite{schulman_proximal_2017,heess_emergence_2017,peng_deeploco_2017}, and eliminates some of the hand engineering that is required by classical controllers. Deep neural networks, however do not provide insight as to where the policies will be stable. Generally, DRL policies train on the domain they are required to operate, either with a single end-to-end policy~\cite{heess_emergence_2017}, or by learning when to switch by interacting with the target environment~\cite{merel_hierarchical_2019}. Developing policies on a real system that requires exposure to a large set of terrain types during training may be difficult. Instead, it is more practical to design separate policies and then determine when to switch. This approach also allows us to embed prior knowledge into controller design (such as a rudimentary walking policy), to improve the learning outcomes of each individual policy~\cite{tidd_guided_2020}.

In this work, we first train individual DRL policies, each capable of traversing a single respective terrain artifact. The RoA of each controller is expanded using curriculum learning, while ensuring there exists a set of states common between policies (RoA overlap). We then train a neural network that estimates the probability of successful switching, given the robot state and target policy. We evaluate our approach on a random sequence of terrain types (an example is shown in Fig~\ref{fig:main_pic}). We show that learning to switch is more stable than heuristic methods. We also compare our method to learning approaches trained on the test domain (sequence of terrain artifacts rather than individual artifacts), and show that we have comparable results despite interacting with only one terrain condition at a time.

The contributions of this work are two-fold:
\begin{itemize}
\setlength\itemsep{0pt}
    \item We train DRL policies for each terrain type following our previous work using Curriculum Learning. We extend this method by investigating the effect of state overlap on safe policy switching
    \item We design a policy switching network that estimates when the robot is in the RoA for the next policy by predicting the probability of success switching in the current state
\end{itemize}

The organization of this paper is as follows. After reviewing the relevant literature in Sec~\ref{sec:related_work}, we describe our method in Sec~\ref{sec:method}. We define the problem of interest in Sec~\ref{secsec:problem}, present how we train DRL policies for individual terrain types with curriculum learning in Sec~\ref{secsec:individual} and describe our switch estimator in Sec~\ref{secsec:switching}. We present our results in \ref{sec:experiments}, before concluding with a brief discussion in Sec~\ref{sec:conclusion}.

\section{Related Work}
\label{sec:related_work}

Controllers developed with bipedal robots with classical methods can perform complex behaviours such as ascending and descending stairs~\cite{ching-long_shih_ascending_1999}, balancing on a Segway~\cite{gong_feedback_2019}, and executing a jump~\cite{xiong_bipedal_2018}. Humanoid robots often employ a set of control primitives, each individually developed and tuned, such as Dynamic Movement Primitives which uses discrete and rhythmic controllers that allow a humanoid robot to play the drums and swing a tennis racket~\cite{schaal_dynamic_2006}. Hauser et al.~\cite{siciliano_using_2008} uses primitives to place and remove a foot from contact, enabling a robot to walk on uneven terrain, up a step, and climb a ladder. Motion capture primitives can be stitched together with Hidden Markov Models to create sequences of motions~\cite{kulic_incremental_2012}. Controllers developed with classical methods usually require extensive human engineering.

Linking a sequence of controllers together by understanding where each works has been demonstrated on a juggling robot, the stable switching of behaviours is known as sequential composition~\cite{burridge_sequential_1999}. A key component is determining the region of attraction (RoA) for each policy, which is defined as the set of states that whether engaging a given policy will converge to a target set of states defined for that policy. Estimating the RoA overlap between controllers can be difficult for robots with high state spaces, however it can be simplified by defining a set of pre and post conditions for each controller~\cite{faloutsos_composable_2001}, or by providing a rule-based bound on parameters like heading angle and switching frequency~\cite{gregg_control_2012}. 
Motahar et al~\cite{motahar_composing_2016} switch between straight, left turn, and right turn controllers operating a 3D simulated biped using a reduced dimension hybrid zero dynamics control law. These methods result in combinations of complex primitives, though they require a mathematically defined RoA, or a hand designed switching criteria for each primitive. RoA expansion can also be considered to provide a greater overlap between controllers~\cite{borno_domain_2017}. Borno et al.~\cite{borno_domain_2017} estimates the RoA for a simulated humanoid using multiple forward dynamics model simulations, a similar approach to ours.

DRL offers an alternative to classical control methods, policies learn how to act by interacting with the environment. DRL methods have demonstrated bipedal walking over complex terrain~\cite{schulman_proximal_2017,heess_emergence_2017,peng_deeploco_2017}, and performing complex maneuvers replicating motion capture~\cite{peng_deepmimic_2018}. Usually DRL methods are limited to solving a single task end-to-end, characterised by a scalar reward function. End-to-end methods require retraining if new terrains are introduced, and in a real setting it may be intractable to train on the complete set of expected conditions.

Designing a single policy that can display multiple behaviours is challenging, and often results in the degraded performance of individual behaviours~\cite{lee_robust_2019}. Combining DRL primitives usually involves training a deep Q network (or a similar discrete switching network) that selects which primitive to use~\cite{liu_learning_2017,merel_hierarchical_2019,lee_robust_2019}, or learning a complex combination of primitives~\cite{peng_mcp_2019}. Other hierarchical approaches learn the primitives and selection network together~\cite{bacon_option-critic_2018}, or with several task-dependent selection networks~\cite{frans_meta_2017}. Training in each of these examples needs access to all expected environment conditions during training, and adding new primitives requires retraining of the selection policy.

While estimating the complete RoA of a policy is possible for systems with a relatively small state space, such as the single~\cite{berkenkamp_safe_2017},~\cite{najafi_learning_2016}, or double inverted pendulum~\cite{randlov_combining_2000}, for more complex systems the RoA is difficult to determine. Where it is possible to differentiate between unmodeled regions of the state from those that are well behaved, DRL can guide the agent back to where a classical controller can take over~\cite{najafi_learning_2016},~\cite{randlov_combining_2000}. Other work with DRL has shown that expanding the RoA of each primitive with transition policies improves switching~\cite{lee_composing_2019}. Designing policies with DRL for complex walking robots and  combining these behaviours with RoA estimation is a promising approach to scale the capabilities of legged robots, and is an area of research that requires more investigation. 

In our work, we pre-train a set of policies with DRL to traverse complex terrain artifacts. We not only ensure there is a RoA overlap between policies, but also estimate when the overlap occurs such that policies can be switched safely, without the need to train on the complete set of terrain combinations.



\section{Method}
\label{sec:method}

\begin{figure}[tb!]
\vspace{2mm}
\centering
\includegraphics[width=0.95\columnwidth]{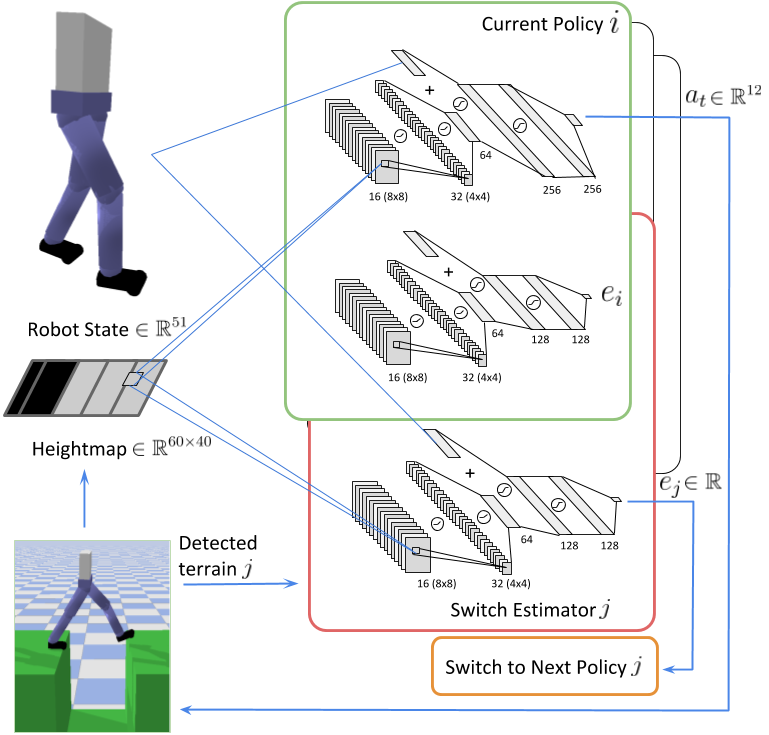}
\caption{
We pre-train a suite of deep reinforcement learning (DRL) policies for a set of terrain conditions, and for each policy we train a switch estimator. Using the current policy $\pi_i$, and detecting upcoming terrain type $j$, we use the switch estimator output $e_j$ to determine when the robot is in a suitable configuration to safely switch to the next policy $\pi_j$. Each policy $\pi_i$ (where $i \in$ $\{$walk, jump, gaps, stairs$\}$) is a neural network with inputs $s_t=\{$robot state, heightmap$\}$ and outputs $a_t$ that are the torques applied to each joint. Each switch estimator $\mathcal{E}$ is a neural network with inputs $s_t=\{$robot state, heightmap$\}$ and outputs $e_j$, where $e_j \in [0,1)$ indicates that the current state is in the region of attraction (RoA) for $\pi_j$: $s_t \in R(\pi_j)$. Switching to $\pi_j$ is safest when $e_j$ is high.
}
\label{fig:method}
\vspace{-3mm}
\end{figure}

A summary of our method is outlined in Fig.~\ref{fig:method}. For each terrain type $i$, where $i \in$ $\{$walk, jump, gaps, stairs$\}$, we train a neural network policy and a switch estimator. We assume access to an oracle terrain detector to determine the upcoming terrain type, and therefore what the next policy will be. The switch estimator $\mathcal{E}_j$ for the next policy $\pi_j$ is activated once the next terrain type is detected, with output $e_j \in [0,1)$ indicating the confidence level that the robot is in the RoA of the next policy.


\subsection{Problem definition}
\label{secsec:problem}
The problem we consider is a biped in a 3D environment, with 12 torque controlled actuators using the Pybullet simulator~\cite{coumans_pybullet_2020}. The task of the biped is to navigate obstacles typically navigated by a human delivery driver, as such the robot must traverse stairs, gaps, and a large step. We consider the state $s_t=[rs_t, I_t]$ to be the robot state $rs_{t}$ and heightmap image $I_t$ at time $t$.


\textbf{Robot state}: $rs_t=[J_t, Jv_t, c_t, c_{t-1}, v_{CoM,t}, \omega_{CoM,t}, \\ \theta_{CoM,t}, \phi_{CoM,t}, h_{CoM,t}]$, where $J_t$ are the joint positions in radians, $Jv_t$ are the joint velocities in rad/s, $c_t$ and $c_{t-1}$ are the current and previous contact information of each foot, respectively (four Boolean contact points per foot, plus one variable for no points in contact), $v_{CoM,t}$ and $\omega_{CoM,t}$ are the linear and angular velocities of the body Centre of Mass (CoM), $\theta_{CoM,t}$ and $\phi_{CoM,t}$ are the pitch and roll angles of the CoM, and $h_{CoM,t}$ is the height of the CoM above the terrain. All angles except joint angles are represented in the world coordinate frame. In total there are $51$ elements to the robot state, which is normalised by subtracting the mean and dividing by the standard deviation for each variable (statistics are collected as an aggregate during training).

\textbf{Heightmap}: Perception is provided in the form of a heightmap that moves with the x,y,z and yaw positions of the robot body. A heightmap is a common perception method used in robotics, usually extracted from range measuring sensors such as depth cameras, laser scanners or stereo cameras~\cite{fankhauser_universal_2016}. Other work in DRL for walking uses perception from state information~\cite{lee_composing_2019}, RGB cameras~\cite{merel_hierarchical_2019}, or a heightmap~\cite{peng_deeploco_2017}. In early experiments, we found that using a heightmap improved policy performance compared to providing the ground truth terrain pose. The height map is scaled from 0 (the CoM of the robot) to 1 (2m below the CoM), and has a resolution of $[60,40]$, with a grid size of $0.025$m. The robot itself does not appear in the heightmap, and is centred with a larger view in front. The effective field of view is $0.9$m in front, $0.6$m behind, and $0.5$m to each side. A depiction of the heightmap is shown in Fig.~\ref{fig:method}.

\subsection{Training Policies for Individual Terrain Types}
\label{secsec:individual}
Our switching method requires policies to have a RoA overlap, and provided this condition is met we can employ policies derived from any method. We choose DRL as our method to populate our set of policies. In this section we first introduce the reinforcement learning problem and the algorithm of choice. We then introduce our reward function, and finally our curriculum learning approach that ensures policies have a RoA overlap. 

\textbf{Deep Reinforcement Learning}: We consider our task to be a Markov Decision Process (MDP) defined by tuple $\{\mathcal{S},\mathcal{A}, R, \mathcal{P}, \gamma\}$ where $s_t \in \mathcal{S}$, $a_t \in \mathcal{A}$, $r_t \in R$ are state, action and reward observed at time $t$, $\mathcal{P}$ is an unknown transition probability from $s_t$ to $s_{t+1}$ taking $a_t$, and $\gamma$ is a discount factor. The reinforcement learning goal is to maximise the sum of future rewards $R = \sum_{t=0}^{T}\gamma^tr_t$, where $r_t$ is provided by the environment at time $t$. For continuous control, actions are sampled from a deep neural network policy $a_t\sim\pi_\theta(s_t)$, where $a_t$ is a torque applied to each joint. We update the weights $\theta$ of the policy using Proximal Policy Optimisation (PPO)~\cite{schulman_proximal_2017}.

\textbf{Guided Curriculum Learning}: Curriculum learning is a systematic way of increasing the difficulty of a task~\cite{bengio_curriculum_2009}, and results in learning attractive walking gaits quickly~\cite{yu_learning_2018}. In our work, \textbf{Guided Curriculum Learning}~\cite{tidd_guided_2020}, we apply curriculum learning to a sequence of tasks, where each stage is completed before the commencement of the next stage. The difficulty of each curriculum is increased as milestones are reached with average episode reward used as the milestone for all stages: as the robot consistently achieves a threshold reward, the difficulty of the current stage is increased. We employ three stages: increasing terrain difficulty while guided by expert forces, reducing the expert forces, and increasing perturbations. 

The first stage is increasing the terrain difficulty with external guidance from a rudimentary target walking policy. We apply external forces to stabilise the CoM (similar to~\cite{yu_learning_2018}), and to each joint. Forces applied to each joint are determined from a PD controller tracking the rudimentary target walking policy. Once the respective terrain is at its most difficult setting (maximum stair height, step height, and gap length), all external forces are annealed to zero. The third stage increases the magnitude of perturbations. We found adding strong perturbations early hindered training, while increasing gradually in the last curriculum stage allowed for greater final disturbances and more robust policies. By increasing the perturbations we also increase the RoA of the respective policy, i.e. the set of states the policy can safely operate within is now larger. The full details on our curriculum learning approach, including the reward used for training each policy, can be found in our previous work~\cite{tidd_guided_2020}. We extend this work by comparing the effect of starting all policies from the same set of initial conditions (a standing pose), compared to starting from a random joint configuration, and thus enforce there to be a set of states that are common between policies (RoA overlap). The policies must pause for a short duration (0.5 seconds) before commencing the location behaviour. 

\begin{figure}[tb!]
\centering
\subfloat[]{%
\includegraphics[height=0.3\columnwidth]{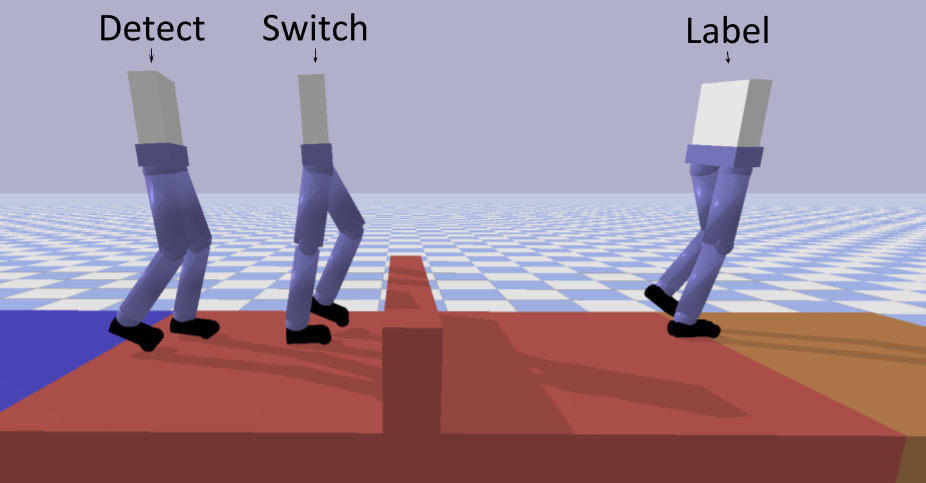}}
\hfill
\subfloat[]{%
\includegraphics[height=0.3\columnwidth]{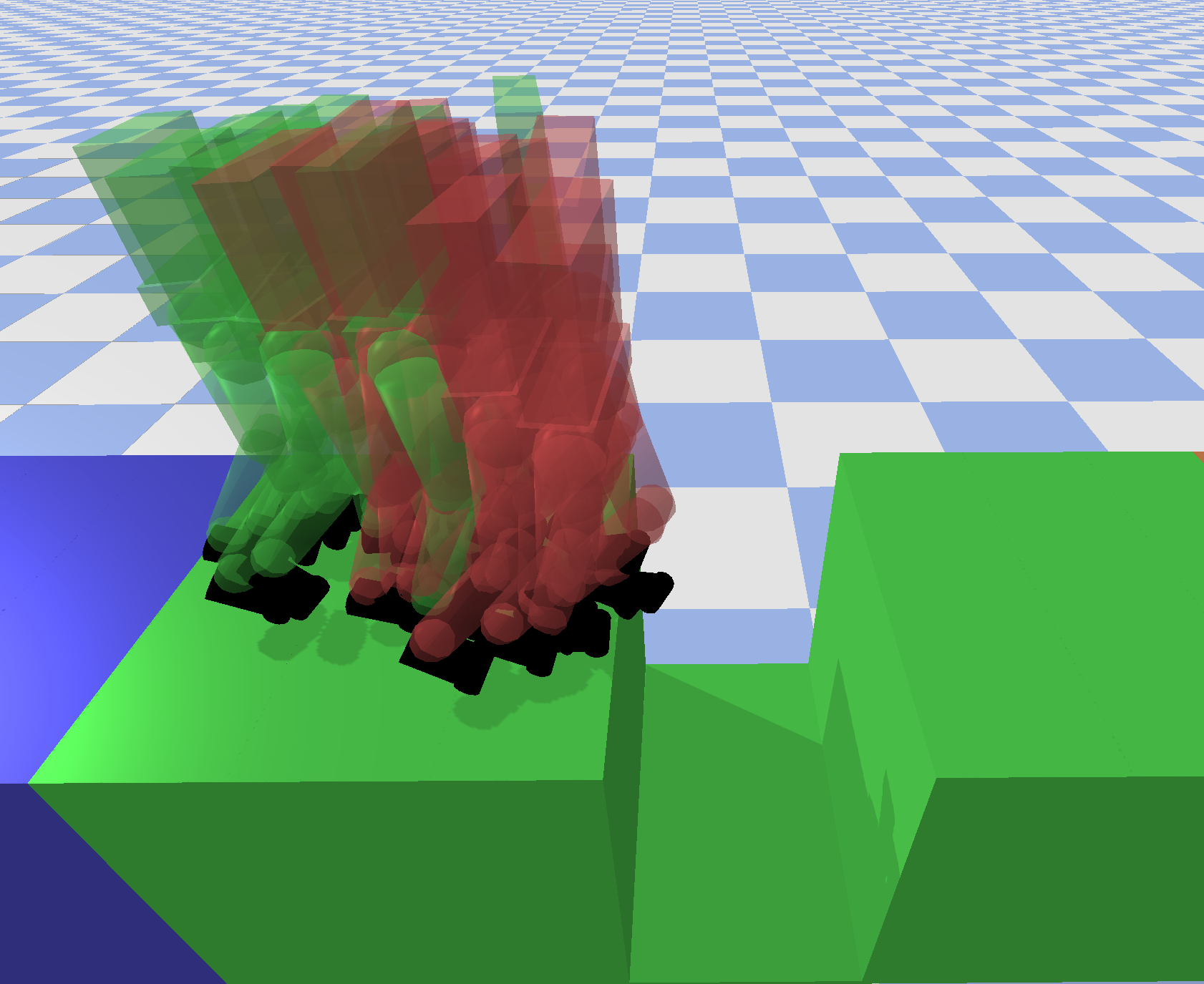}}
\hfill
\caption{a) A switch estimator is trained for each policy, with data collected from each terrain type separately. Data points are collected around the switch point and given a label of 1 if the robot reaches the goal state, 0 otherwise. b) Early switching results in successful transitions for some policies (green), whereas switching late is more likely to fail (red).}
\label{fig:estimator}
\vspace{-3mm}
\end{figure}

\subsection{Learning When to Switch}
\label{secsec:switching}
Switching from one policy to another by estimating when a state is in the RoA for the next policy is an idea that stems from the sequential composition framework~\cite{burridge_sequential_1999}. Each policy $\pi_i$ has a goal set $\mathcal{G}(\pi_i) = \{s_i^*\}$ where $s=[rs, I]$, and a Region of Attraction (RoA) $\mathcal{R}(\pi_i) = \{s_0 \in \mathcal{S} : \lim_{t\to\infty} s(t,s_0) = \mathcal{G}(\pi_i)\}$. $\mathcal{G}(\pi_i)$ is the set of states policy $\pi_i$ was designed to operate within (Goal or Invariant Set), and $\mathcal{R}(\pi_i)$ is the set of states that will converge to a state in $\mathcal{G}(\pi_i)$. 

The complete set of states of the RoA for a complex biped is difficult to determine. Instead, we estimate when the robot is in the RoA of a given policy by training a neural network switch estimator. We train a switch estimator for each policy. In other words, each policy has its own switch estimator that is activated once its respective target terrain type is detected. If we switch policies when the switch estimator for the upcoming policy predicts a high value, then the likelihood of the robot safely traversing the target terrain is increased.

The data collection procedure for switching is as follows. A sample is collected by running $\pi_j$ on terrain condition $i$ and switching to $\pi_i$ a random distance from the terrain artifact (where $j \neq i$). As shown in Fig.~\ref{fig:estimator}a), there are three steps to data collection that occur for each sample: detection, switching, and labelling. Detection occurs when the terrain artifact appears $0.9$m from the robot body. We switch to the target policy at a distance $\sim\mathcal{U}(0,0.9)$m from the artifact. A total of 5 data points (robot state + heightmap) are collected for each sample, including the data point when switching occurred at timestep $t$ and the surrounding data points: $[t-2, t-1, t, t+1, t+2]$. The sample (all 5 data points) is labelled as successful (1) if the robot is able to traverse the terrain artifact and remain stable for two complete steps past the terrain. Otherwise, all data points in the sample are labelled as failed (0).  



Each switch estimator $\mathcal{E}_i$ is trained to minimise the loss $L(\phi) = (\mathcal{E}_i(s_t) - y_t)^2$, where $\phi$ is the neural network weights, $s_t$ is the state and $y_t$ is the corresponding label for data point $t$. The final layer of the network is a single output with a sigmoid activation. Data is split into 90\% training and 10\% validation, with a 50/50 split of positive and negative labels forced during training and validation. We collected a total of 100000 data points for each terrain type, and trained each switch estimator for 1500 epochs.

\section{Experiments}
\label{sec:experiments}

In this section, we present the results of our simulation experiments. First, in Sec~\ref{sec:single} we analyze the performance of each policy on the single artifact type they are trained on. Second, in Sec~\ref{sec:multi} we present results on terrains that have a sequence of randomly ordered multiple artifact types. In Sec~\ref{sec:comparison} we discuss how our approach compares with learning methods that are trained directly on the multi-terrains. For performance evaluation, we measure two metrics: 
\begin{itemize}
\setlength\itemsep{0pt}
    \item \textbf{\% Total Distance:} How far the robot has travelled as a percentage of the total terrain length.
    \item \textbf{\% Success:} The percentage of cases where the robot was able to get to the end of the terrain without falling.
\end{itemize}

Each terrain artifact is made up of horizontally stacked boxes, with each trial composed of 7 artifacts. In the single-terrain case all artifacts are of the same type, whereas in the multi-terrain case they are randomly selected among all terrain types. Terrain artifacts consist of Stairs (up and down) with a rise of 0.17m, Gaps that are 0.7m in length, and Steps 0.3m high and 0.16m in length (the size of each obstacle is kept fixed). The terrain width is sampled uniformly from $\mathcal{U}(1.1,1.7)$m and fixed for each trial, flat segments (including the run of stairs) have a length sampled uniformly from $\mathcal{U}(0.36,0.44)$m. We do not consider the efficacy of policies on terrains outside of these ranges, some discussion on out of distribution operation is presented in~\cite{tidd_guided_2020}. 

\subsection{Single Artifact Type Terrains}
\label{sec:single}

Table~\ref{tab:individual_results} shows the percentage of total track distance covered and the success rate for each terrain type. Each policy is evaluated for 500 trials with 7 artifacts per trial. In these results no switching occurs. We evaluate two sets of policies according to the initial conditions: if the robot started with a standing joint configuration and was required to pause for a short duration before walking, or if the robot started with a random joint configuration with no pause. Of the policies starting from a standing joint configuration, the Jumps policy performed best, with a success rate of 87.8\%. Both conditions for starting state were effective at learning policies, though we observed that random initial joint positions resulted in more successful policies for the Gaps and Stair artifacts (75.5\% and 59.0\%). We suggest this improvement may be the result of imposing fewer state limitations on the emergent behaviours. The percentage of the total distance metric was consistent with the success rates.



\begin{table}[]
    \vspace{3mm}
    \centering

    \begin{tabular}{M{0.9cm}|M{1.65cm}|M{1.5cm}|M{1.65cm}|M{1.5cm}}
         
         Policy & \% Total Dist. &\% Success & \% Total Dist. rand & \% Success rand\\
         \hline
         Jumps  & 93.1 & 87.8 & 86.2 & 77.5\\
         Gaps   & 76.0 & 56.3  & 87.4 & 75.5 \\
         Stairs & 71.4 & 52.1  & 78.7 & 59.0 \\

         
         \hline
    \end{tabular}
    \caption{Results of each individual policy on its corresponding terrain type. \% of total track distance covered and \% success is shown. We also report the success of policies that are trained with random initial joint positions.}
    \label{tab:individual_results}
\end{table}


\subsection{Multiple Artifact Type Terrains}
\label{sec:multi}
Our switching method is compared with several other methods, with results displayed in Table~\ref{tab:test_results}. All tests are performed with 2500 trials of a test world consisting of 7 terrain artifacts randomly selected from gaps, jumps and stairs. Each trial begins with a flat segment and the walk policy. Artifacts are restricted so there may not be two sequential artifacts of the same type. For the combination test worlds, only widths are fixed for an episode, each box length is randomly chosen. The number of boxes in each artifact varies from 4 to 9, while ensuring at least 0.9m of flat surface before each artifact to be consistent with how our switch estimator data is collected.


We compare our policy switching method with several heuristic methods, with the only difference between the methods being the timing of the switching.
\begin{itemize}[leftmargin=*]
    \item \textbf{Random:} Switching occurs a random distance after the artifact is detected.
    \item \textbf{On detection:} Switching occurs as soon as a next artifact is detected.
    \item \textbf{Lookup table:} We evaluate the optimal distance to switch for all switch combinations. On each artifact separately, we record the success rate of switching from any other policy to the target policy from 0 to 0.9m from the artifact (0.01m resolution).  Therefore, each switchable combination (e.g. $\pi_j$ to $\pi_i$) has a table of size $90$. We then take the $\arg\max$ of each permutation to determine the optimal switch distance for each artifact given the current policy and the target policy.
    \item \textbf{Centre of mass over feet:} Having the CoM over the support polygon created by the grounded foot is a well known  criteria for the stability of legged robots~\cite{faloutsos_composable_2001}. We switch when the following two conditions are true: 1) CoM is over the support polygon and 2) Both feet are within a tolerance distance to each other.
    \item \textbf{Our switch estimator:} We predict the probability of success for switching in the current state with our switch estimators. We threshold our predictions at 0.85 which empirically gave the best overall success for transitions. If terrain condition $i$ is detected and $\mathcal{E}_i(s_t) \geqslant 0.85$ we switch to $\pi_i$. All policies are trained by starting with the same initial state (a standing configuration), followed by a small pause before the robot attempts the respective terrain.
    \item \textbf{Our switch estimator without region of attraction overlap:} We test our method using policies trained without considering a region of attraction overlap. The robot is initialised from a random set of joint positions, and is not required to pause before commencing locomotion. All other training and terrain properties are consistent with our switch estimator method, and all policies achieve comparable, or better success rates than policies forced to start from a standing initial state (Table.~\ref{tab:individual_results}).
\end{itemize}

For \textbf{CoM over feet} and our \textbf{Switch Estimator} methods, if there is no state where the condition to switch is met, then we force-switch to next policy when the robot is within 1cm of the corresponding artifact (other methods switch at a specified distance to the artifact).

\begin{table}[]
    \vspace{3mm}
    \centering
    \begin{tabular}{c|c|c|c|c|c}
         \hline
         Switch Method & \scriptsize{\% Total Dist.} & \scriptsize{\% Success} & \scriptsize{\% Gap Fail} & \scriptsize{\% Jump Fail} & \scriptsize{\% Stair Fail}\\   
         \hline
         Random  & 42.7 & 10.1 & 70.6 & 11.6 & \textbf{7.6} \\
         On detection  &   75.6 & 60.1 & 21.4 & 3.3 & 15.2 \\
         Lookup table &   76.3  & 59.0  &  21.2 & 4.7 & 14.9\\
         CoM over feet  & 79.1 & 66.8 &  17.6 & 2.7 & 12.9 \\
         Ours & \textbf{82.4} & \textbf{71.4} & \textbf{16.7} & \textbf{2.4} & 9.4\\   
         Ours no overlap& 17.5 & 0.7 &  58.6 & 12.9 & 27.7 \\   
         \hline
         *DQN  &  83.2 & 72.4 &  15.1 & 3.7 & 8.8  \\
         *End-to-end  &  \textbf{92.7} & \textbf{90.2} & \textbf{4.2} & \textbf{1.8} & \textbf{3.7}\\
         \hline
    \end{tabular}
    \caption{Results of each switching method on multi-terrain test worlds (2500 trials). The \% of the total distance covered, and the success rate for the robot reaching the end of the trial is shown. The final three columns show the \% of failures for each artifact type.\\ *Trained on the test domain (combinations of terrain types).}
    \label{tab:test_results}
\end{table}

Table~\ref{tab:test_results} shows the results for multi-terrain test worlds. For less dynamic platforms it may be possible to switch controllers at any time after obstacle detection, however that was not the case for our scenarios: \textbf{Random} switching performed the worst as expected with only 10.1\% success rate. Switching as soon as an artifact is detected does better than randomly switching with 60.1\% success, we suggest this is because switching early might give the policies enough time to align to an artifact. Fig.~\ref{fig:estimator}.b) shows it is preferential to switch early for some policies. \textbf{Lookup table} method performed similarly to the \textbf{On Detection} method. \textbf{CoM over feet} method performed relatively well with 66.8\% success. An advantage of this method is that it does not require further data collection once individual policies are trained, providing insight into where RoA overlap occurs for most policies.  \textbf{Our} switching method was the most robust, whilst following our condition that we only have access to individual terrain artifact at training time, traversing 82.4\% of the terrain and reaching the final zone in 71.4\% of trials. Finally, we show the effect of training policies without considering RoA overlap. \textbf{Ours no overlap} performed poorly. Despite the capabilities of the individual behaviours reported in Table.~\ref{tab:individual_results}, policies trained without explicit state overlap are unable to successfully complete the multi terrain test environment. This result highlights the importance of RoA consideration when training dynamic policies where switching between controllers may be required. 

\begin{figure}[tb!]
\vspace{2mm}
\centering
\includegraphics[width=0.95\columnwidth]{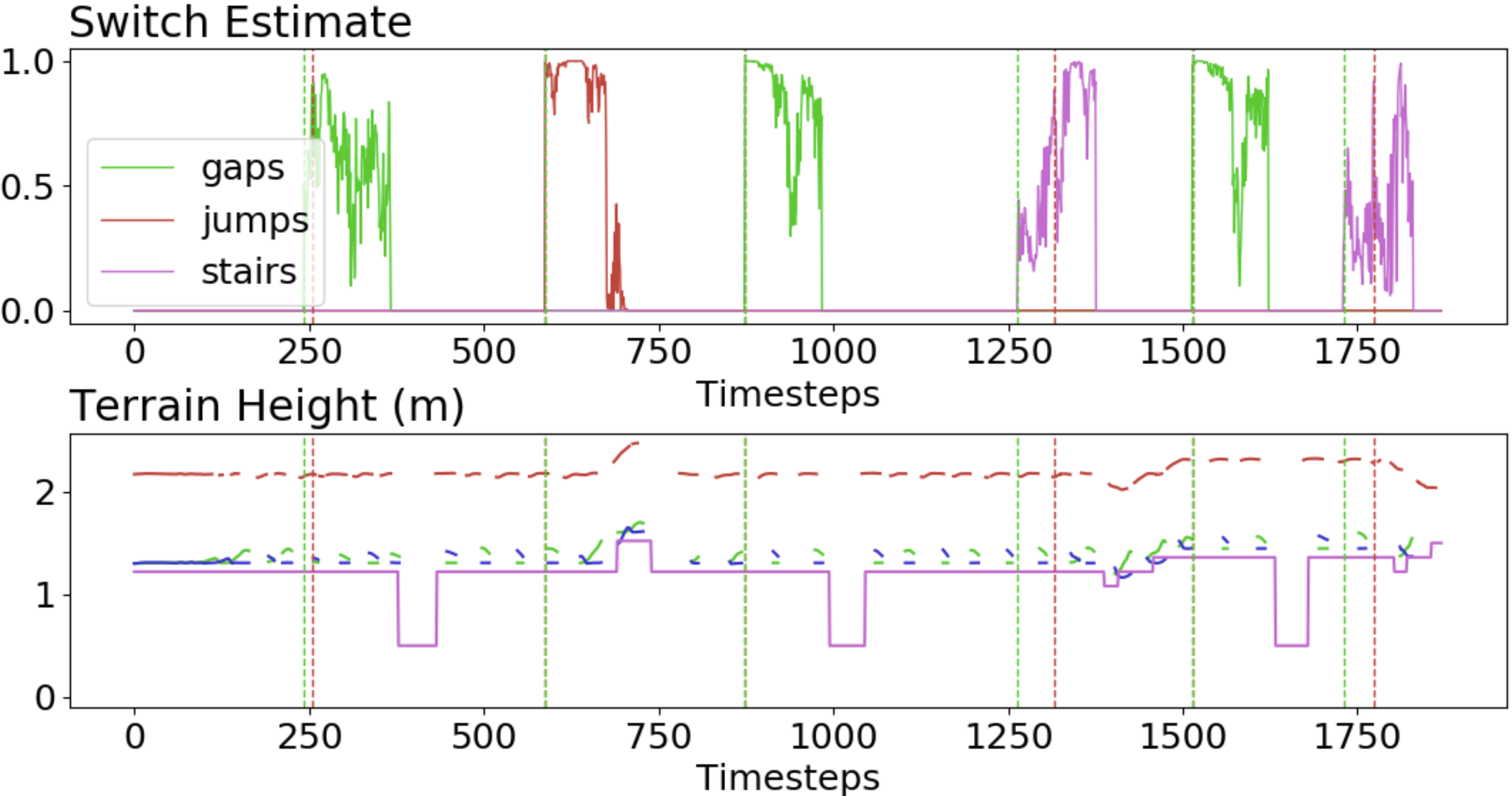}
\caption{Our method on a single trial in the test world. The top figure shows the switch estimate from the time the next artifact is detected until the robot has past the artifact. The bottom figure shows the terrain (pink) with foot (blue and green) and CoM height (red) when the CoM is over the stance foot. Green vertical lines indicate when the terrain was detected, red vertical lines show when switching occurred.}
\label{fig:plot}
\end{figure}

Fig.~\ref{fig:plot} shows a single trial of the test world using our method. We can see the switch estimator predictions for each policy in the top figure, and the times our predictions prevented early switching (vertical green lines show detection, vertical red lines show when switching occurs). We can also see in the bottom plot the times that the CoM is over the feet, and instances our estimator switched outside of these regions ($\text{timestep}\approx875$).

The final three columns of Table~\ref{tab:test_results} shows the failure distribution of each switch method. We record where the robot was when failure occurs, given as a percentage of all trials. For each method, the most failures happened after switching to the Gaps policy. For instance, it contributed to to 16.7\% of a total (28.6\%) failures for our switching method. For all terrain types, these results show less failures than the single terrain experiments reported in Table~\ref{tab:individual_results}. This suggests we are improving the overall robustness of our behaviour suite using our switch estimator.

\subsection{Comparison with training on multiple artifact terrains}
\label{sec:comparison}

For completeness, we also look at other methods that involve training on a full combination of terrain artifacts, including training a \textbf{Deep Q Network (DQN)} using our pre-trained policies, and a single \textbf{end-to-end policy}. Results are displayed in the last two rows of Table.~\ref{tab:test_results}. The \textbf{DQN} was provided with the artifact detection in the form of a one hot encoding appended to the robot state, and received the same reward as each policy, with an additional reward provided by our oracle terrain detector encouraging the activation of the correct policy. We note that without the additional reward provided by the oracle the \textbf{DQN} failed to learn. For the \textbf{end-to-end} learning method, we applied the same reward and curriculum learning approach (from~\cite{tidd_guided_2020}) as our policies. Robot state + heightmap is provided as input, and torques are generated at the output. We note that without the curriculum the \textbf{end-to-end} method failed to learn.

Our estimator performs similarly to the \textbf{DQN} despite not seeing more than one terrain artifact type during training (71.4\% success compared to 72.4\%). While learning was substantially longer for the \textbf{end-to-end} policy (approx. two times longer than an individual policy), it performed the best with 90.2\% success rate. While these methods trained on the test domain perform well, it may not be possible to train on all expected terrain conditions on a real platform, where training policies separately allows us to refine each policy in a constrained setting. 


We consider the case where a new terrain is introduced to an existing behaviour suite with two behaviours. With our method, we train a policy for the new terrain obstacle, and retrain the switch estimator for each policy (three in total) to include the new behaviour. We record the wall clock time for training a single policy and the respective switch estimators, and compare to the time to retrain the \textbf{DQN} and \textbf{end-to-end} methods. It takes approximately 17 hours to train a single policy, and an hour to collect data and train a single switch estimator. The \textbf{DQN} method takes about 8 hours to train (after the new policy is available), and the \textbf{end-to-end} method takes roughly 34 hours to train. Table.\ref{tab:re_training} shows an estimate of wall clock time for each method when a new obstacle is introduced. It can be seen that our method is the most efficient to incorporate new behaviours. Furthermore, our method has the same training complexity as new behaviours become available, where the training becomes more difficult for the \textbf{DQN} and \textbf{end-to-end} methods as more diverse obstacles are introduced. A current limitation of our method is that we collect switching data from all available behaviours, requiring retraining of all switch estimators with each new behaviour. This could be improved by collecting data from a common walking policy, where only a single switch estimator would be trained with a new behaviour. We leave this improvement for future work.


\begin{table}[]
    \vspace{3mm}
    \centering
    \begin{tabular}{c|c|c}
         \hline
         \scriptsize{\textbf{DQN (hrs)}} & \scriptsize{\textbf{end-to-end (hrs)}} & \scriptsize{\textbf{Ours (hrs)}}\\   
         \hline
         25 & 34 & \textbf{20} \\
         \hline
    \end{tabular}
    \caption{Wall clock time for retraining when a new terrain is introduced. Each method is trained with the same number of CPU cores of the same architecture.}
    \label{tab:re_training}
\end{table}

\section{Conclusion}
\label{sec:conclusion}
We introduce a novel method for estimating when to switch between a set a of pre-trained policies. We show that our method improves the stability of switching compared to heuristic methods, where policies and switch estimators only have access to a single terrain during training. Our method also performs comparable to a DQN trained on test conditions (all terrain types), which may not be possible for real systems. By designing policies separately we are able to refine controllers in a constrained setting, and embed prior knowledge about the the required behaviour. Similar to a human delivery driver, our bipedal agent can traverse several difficult terrain obstacles. By understanding when to switch between behaviours our method can scale to any number of terrain conditions, where methods trained on the test set of terrains require more costly retraining (DQN and end-to-end).

We design our policies with an overlapping Region of Attraction (RoA), and show that this is required for success transitions between behaviours. However, this may not always be the possible, particularly for specialist policies where the RoA represents a very narrow set of states that do not overlap with a simple walking policy. Future work will look at evoking a setup policy that expands the RoA of a given policy to ensure this overlap exists. A limitation of our method is the assumption that we have a terrain oracle, learning to identify the upcoming terrain type is left for future work. We believe that harnessing the capabilities of DRL and combining controllers in a modular way will allow us to expand the locomotion capabilities of legged robots.

\addtolength{\textheight}{-2cm}   
\bibliographystyle{IEEEtran}
\bibliography{additional_refs}
\end{document}